\pdfoutput=1
\documentclass[11pt]{article}

\usepackage[final]{acl}
\usepackage{times}
\usepackage{latexsym}
\usepackage[T1]{fontenc}

\usepackage[utf8]{inputenc}

\usepackage{microtype}

\usepackage{algorithmic} 
\usepackage{amsmath}

\usepackage{mathtools}
\usepackage{physics}
\usepackage{verbatim} 
\usepackage{url}
\usepackage{todonotes}
\usepackage{dirtytalk}
\usepackage{algorithm}
\usepackage{algorithmic}
\usepackage{booktabs}
\usepackage{multirow}
\usepackage{adjustbox}
\usepackage{changepage}
\usepackage{caption}
\usepackage{subcaption}
\usepackage{bm}
\usepackage{lipsum}
\usepackage{breqn}
\usepackage{array}
\title{Cost-Effective Training in Low-Resource Neural Machine Translation}

 \author{
        Sai Koneru, Danni Liu and Jan Niehues\\
        Department of Data Science and Knowledge Engineering, Maastricht University \\ 
        \texttt{s.koneru@student.maastrichtuniversity.nl}\\
        \texttt{\{danni.liu,jan.niehues\}@maastrichtuniversity.nl}
        }

\begin{document}

\maketitle

\begin{abstract}
While Active Learning (AL) techniques are explored in Neural Machine Translation (NMT), only a few works focus on tackling low annotation budgets where a limited number of sentences can get translated. Such situations are especially challenging and can occur for endangered languages with few human annotators or having cost constraints to label large amounts of data. Although AL is shown to be helpful with large budgets, it is not enough to build high-quality translation systems in these low-resource conditions. In this work, we propose a cost-effective training procedure to increase the performance of NMT models utilizing a small number of annotated sentences and dictionary entries. Our method leverages monolingual data with self-supervised objectives and a small-scale, inexpensive dictionary for additional supervision to initialize the NMT model before applying AL. We show that improving the model using a combination of these knowledge sources is essential to exploit AL strategies and increase gains in low-resource conditions. We also present a novel AL strategy inspired by domain adaptation for NMT and show that it is effective for low budgets. We propose a new hybrid data-driven approach, which samples sentences that are diverse from the labelled data and also most similar to unlabelled data. Finally, we show that initializing the NMT model and further using our AL strategy can achieve gains of up to $13$ BLEU compared to conventional AL methods.





\end{abstract}

\section{Introduction}
There are several thousand languages in today's world, with millions of people knowing only their native language. This creates a language barrier and is a hindrance to communication in this globalized world. Translation technologies are essential to overcome the language barries and enable communication between monolingual speakers. Neural Machine Translation (NMT) systems \cite{bahdanau2015neural, vaswani2017attention} have significantly advanced translation quality to alleviate this problem. Supervised NMT models rely on vast amounts of parallel sentences to translate between languages with high quality. But, the labelled data is not available for many language pairs. 


Unsupervised NMT (UNMT) \cite{lample2018unsupervised,artetxe2018unsupervised} and  UNMT with multilingual transfer \cite{fraser-2020-findings,garcia2021harnessing,li-etal-2020-reference} are promising research directions to tackle this problem. The former learns to translate, relying on monolingual corpora but fails in practical conditions when dealing with distant low-resource language pairs \cite{kim2020and,marchisio2020does}. The latter approach uses parallel data between similar high-resource language pairs and generates decent quality. However, it is not enough to produce high-quality translations for several language pairs in both directions (source $\leftrightarrow$ target). Labelled data between the language pair in focus is necessary to attain SOTA performance.

However, human annotation of sentences poses several challenges: 1) Costly and time-taking; 2) Bilingual translators for several language pairs are hard to find. Hence, annotating large amounts of parallel sentences for low-resource languages is impractical and expensive. We need to design a training procedure that is cost-effective but also enables the model to translate with adequate quality.

One way to save costs is by employing Active Learning (AL) strategies with NMT \cite{zeng2019empirical,ambati2012active,haffari-etal-2009-active}. The goal of AL is to maximise translation quality for an annotation budget of labelling $B$ sentences. We label only the most informative $B$ sentences in the whole unlabelled dataset using \textit{selection strategies}. Previous works on AL \cite{zeng2019empirical,peris-casacuberta-2018-active} consider annotation budgets between hundred thousand to million sentences. But, it is not always possible to afford the annotation such amounts of data for low-resource languages. Also, current AL frameworks do not utilize the monolingual data which does not require any labelling. Analysis on AL for low-annotation budgets\footnote{We consider budgets that can annotate between $0$ to $50k$ sentence pairs as low-annotation budgets} with exploiting monolingual data is necessary to build good quality NMT systems in realistic scenarios. 


Another way to improve the model without spending significant money is by integrating small, inexpensive bilingual dictionaries. Word translations are compact, can cover different domains and are a cheaper knowledge source to annotate. Exploiting this additional information with monolingual data and combining it with AL can further improve the performance of the model. However, our methods should be robust and be able to utilize smaller dictionaries. 


In this work, we address the challenges above by the following contributions: 

\begin{itemize}
    \item We show that improving the model's quality by pretraining is necessary before applying AL strategies with low annotation budgets. (Table \ref{tab:strategies})
    
    \item We present a novel "\textit{Cross-entropy difference}" selection strategy for AL that is effective in low-resource scenarios. ($\S$ \ref{cediff})
    
    \item We propose a inexpensive pretraining procedure by incorporating a small dictionary (1146 entries) and show that combining this with AL can increase the translation quality up to $13$ BLEU. (Table \ref{tab:strategies})
\end{itemize}

\section{Background: Active Learning in NMT }
\label{Background}
There are several language pairs for which parallel data is hardly available.  To build NMT systems for these languages, we need to create bi-texts by annotating the unlabelled data. Given an annotation budget, we can only afford to label a certain amount of sentences in unlabelled data. However, choosing data points randomly might include annotating uninformative data and incur a waste of resources. 

AL is an effective solution to reduce the amount of labelling. It uses selection strategies\footnote{We follow the terminology in \citet{zeng2019empirical}} ($\psi$) to mitigate this problem. $\psi(\cdot)$ is simply a scoring function to estimate the "importance" of each sentence of the unlabelled data. Choosing the top-scoring sentences can help in maximising the translation quality for an annotation budget. It can use any of the following as input: 1) Labelled data\footnote{We generate the initial labelled data by annotating random batch of sentences.} ($\mathcal{L}$) 2) Unlabelled source data ($\mathcal{U}_{S}$) 3) Batch size ($\mathcal{B}$) 4) Model ($\mathcal{M}$) trained on the available data.

One paradigm is to use the model $\mathcal{M}$ to score each sentence in the unlabelled data. They are grouped as \textit{model-driven} strategies. The key idea is to determine sentences in $\mathcal{U}_{S}$ for which the model is relatively weaker. Round-trip-translation-likelihood (RTTL) \cite{zeng2019empirical,haffari-etal-2009-active} is the current SOTA approach for model-driven strategies. It gives higher score to sentences for which, the model is unsure during back-translation. We generate a intermediate translation $\hat{t}$ for a sentence $s$. Then, we take the average of the log-probability at token level giving $\hat{t}$ as input and asking to reconstruct $s$ at the output. Higher value indicates that the model is more confident and hence $s$ obtains a lower score.



Another paradigm is to compare each sentence $s$ in unlabelled data to the labelled data $\mathcal{L}$ or the whole unlabelled source data $\:\mathcal{U}_{S}$ itself. These methods can be called as \textit{data-driven} strategies. They rely on the following heuristics: 
\begin{itemize}
    \item \textbf{Diversity}: Sampling sentences that are diverse from the existing labelled data $\mathcal{L}$ is important.
    \item \textbf{Density}: The test set follows the same distribution as the unlabelled data. Hence, sampling from dense regions of unlabelled data $\mathcal{U}_{S}$ is beneficial.
    \item \textbf{Hybrid}: Accounting to both of the above metrics with a trade-off.
\end{itemize}

"$n$-gram overlap" \cite{eck2005low} is simple yet an effective data-driven strategy. It only accounts for the diversity metric. Sentences in the unlabelled data $\mathcal{U}_{S}$ are given a higher score, if they have more number of n-grams that are not present in the labelled data $\mathcal{L}$.





\section{Cost-Effective Training in NMT}

\begin{figure*}[ht]
  \includegraphics[width=\textwidth]{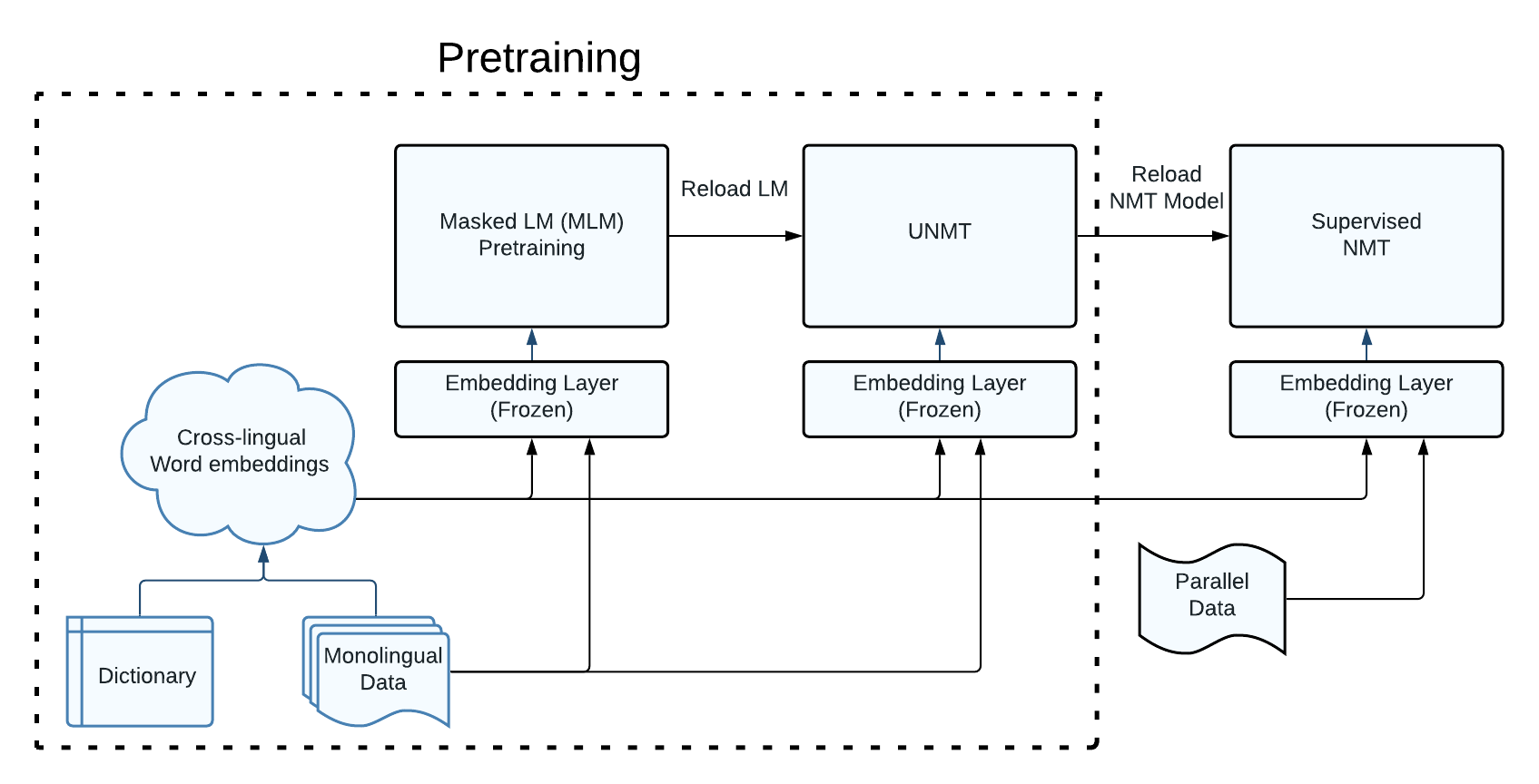}
  \caption{Proposed pretraining procedure to integrate CLWE. The black dotted box denotes the pretraining stage that only uses monolingual data. upervised NMT denotes the fine-tuning phase where we use the parallel data}
  \label{fig:proposedpre}
\end{figure*}

We design a sequence of training steps to exploit additional inexpensive data sources with AL to increase translation quality. The overview of the process is illustrated in Figure \ref{fig:proposedpre}. We utilize the dictionary and monolingual data by training a UNMT system to improve the model. Then, we apply AL to sample informative data and maximise gains. 

As a first step, we use the bilingual dictionary to provide supplementary supervision signal by constructing cross-lingual word embeddings (CLWE) ($\S$ \ref{clwe sub}). We extract embeddings from monolingual data \cite{bojanowski2017enriching} and map them into common space using a small dictionary \cite{artetxe2017learning}. We hypothesize this is useful for supervised NMT in low-resource conditions.

For the second step, we use the monolingual data with CLWE to provide a strong initialization for the NMT model ($\S$ \ref{proposedpre}). We leverage Masked Language modelling (MLM) \cite{devlin-etal-2019-bert} and UNMT \cite{lample2018unsupervised,artetxe2018unsupervised} objectives (self-supervised) on monolingual data to provide a better initialization for the NMT model without the need of annotation. While training on these objectives, we reload the embedding layer with CLWE created in the first step and freeze them for the entire process to always provide cross-lingual signal \cite{banerjee2021crosslingual}.

The last step is to employ AL for labelling and prioritize the annotation of the most informative sentences. We present a novel AL strategy "Cross-entropy difference" that is effective in these low-resource conditions ($\S$ \ref{cediff}). We reload the model trained using self-supervised objectives above as initialization before fine-tuning on the sampled parallel data using AL to achieve higher performance.



\subsection{Integrating Dictionaries}
\label{clwe sub}

\begin{table*}[t]
\begin{adjustbox}{width=410pt,center}
\begin{tabular}{|c|c|}
\hline
\begin{tabular}[c]{@{}c@{}}Dictionary Words\end{tabular} & [\textbf{tomorrow}, \textbf{training}, center]                                                    \\ \hline
Source sentence                                                     & Academic Skills center will focus on \textbf{training} \textbf{tomorrow}       \\ \hline
BPE                                                   & Academic S@@ kills center will focus on train@@ ing tom@@ orrow \\ \hline
\multirow{2}{*}{\begin{tabular}[c]{@{}c@{}}DP-BPE\end{tabular}} & Academic S@@ kills center will focus on \textbf{training} \textbf{tomorrow} \\ \cline{2-2} 
                                                                    & Academic S@@ kills center will focus on train@@ ing tom@@ orrow \\ \hline
\end{tabular}
\end{adjustbox}
\caption{Example for DP-BPE. Words highlighted in bold indicate rare words present in dictionary that get split into multiple words after BPE. The two new encoded sentences after applying DP-BPE are added to the training data.}
\label{tab:jointbpe}
\end{table*}


Incorporating word-to-word translations can increase the potential of NMT models to handle a wider range of words, especially in low-resource conditions. We propose to take advantage of a small dictionary by learning CLWE and utilizing them for low-resource NMT. These embeddings can help in building generalised and cross-lingual NMT models which might be particularly useful in our setup. We can learn the mapping between the monolingual embeddings using a dictionary to create CLWE. Then, we can integrate them with the embedding layer of our NMT model. The only constraint is that the dictionary should contain single token-token entries. But, the current NMT models operate on sub-words using Byte-pair encoding (BPE) \cite{sennrich2016neural}. This is a problem when learning CLWE from the dictionary. Entries consisting of translating rare words would split into multiple tokens. Discarding these (particularly informative) entries would lead to losing information about the mapping between infrequent words.

We can include the infrequent words by simply operating on the word level data. However, this leads to losing all the advantages of operating with sub-words. \citet{chronopoulou-etal-2021-improving} has shown that CLWE is beneficial for UNMT even on sub-word level data. Therefore, we propose a modification to standard BPE in order to retain advantages operating on both word and sub-word tokens. We explain this technique below and denote it as \textit{Dictionary-preserving BPE} (DP-BPE)

First, we create a list of all the words that are present in the dictionary. We consider all the words in the list that will get split into multiple tokens as rare words. Next, we apply standard BPE for sentences that do not contain these rare words. For the remaining text that consists of rare words, we perform the following operations on each sentence:
\begin{enumerate}
    \item Apply BPE on tokens that are not rare words. So, the rare words remain as single tokens.
    \item Apply BPE on all the tokens including the rare words. In this case, these words get split into multiple tokens
    \item Add the above two BPE processed sentences to the existing data. 
\end{enumerate}

We illustrate this process with an example in Table \ref{tab:jointbpe}. The word "tomorrow" and "training" are rare words available in the dictionary which would split into different sub-words. We create two different sentences with selectively applying BPE. We ignore the rare words while applying BPE to form the first sentence. We create another sentence by applying BPE with including the rare words. Finally, we join these two sentences to our dataset.  

There is no alignment between the texts for monolingual data. However, parallel data is aligned between the source and target sentences. The rare words might occur only in the source or only in target or in both sentences. Here, we simply apply standard BPE and DP-BPE at a time and create two new sentence pairs.


Training on the new dataset will result in both the rare word and corresponding sub-words to have similar representation. The word/sub-words will appear in the same context and eventually be treated similarly by the model. Therefore, applying DP-BPE allows us to integrate CLWE with retaining advantages from sub-word based NMT models.

After pre-processing the monolingual and parallel data using DP-BPE, we can start creating CLWE. First, we create the sub-word monolingual embeddings for both languages using a \textit{fasttext} \cite{bojanowski2017enriching} on the monolingual data. 

Next, we align the monolingual embeddings using all the words in the dictionary to build CLWE. As we want to minimize the costs, we only assume having a small dictionary. Hence, we use a \textit{semi-supervised} learning algorithm that is robust to small dictionaries and map the embeddings in a common space using \textit{VecMap} \cite{artetxe2017learning}. Therefore, we are able to build CLWE without spending large amounts on collecting dictionaries.



\subsection{Exploiting Monolingual Data}
\label{proposedpre}

Pretraining in low-resource conditions has been shown to improve the models quality significantly \cite{conneau2019cross,liu2020multilingual}. Therefore, we propose to use the monolingual data to improve the models performance in these challenging conditions. Moreover, having a better model increases its ability to exploit both model and data-driven AL strategies. It is easier for the model to learn from the data selected through various heuristics. Especially, the model-driven strategies need the model to be good enough to accurately identify and learn from data points where it is weak.


We extend the process in \citet{chronopoulou-etal-2021-improving} by integrating dictionaries and use that as a initialization before fine-tuning with AL. We begin by training the encoder using the Masked Language Model (MLM) \cite{devlin-etal-2019-bert} objective on monolingual data for both languages. We build this cross-lingual language model to promote cross-lingual contextual representations. Then, we use this language model for initializing the encoder and decoder and train a UNMT \cite{lample2018unsupervised,artetxe2018unsupervised} system. The UNMT training consists of Denoising auto-encoding \cite{vincent2008extracting} and on-the-fly back translation \cite{sennrich2016improving}. Although this system often struggles to translate between distant languages adequately \cite{kim2020and,koneru2021unsupervised}, it provides a good initialization for the cross-attention and the decoder for fine-tuning.

After training the model as described above, we can start the AL process to select samples. Then, we can fine-tune the model developed using monolingual data on the chosen data points.

\subsection{Effective Sampling for fine-tuning}
\label{cediff}

Model-driven strategies depend on the model to estimate where it is weak. However, in low-resource conditions, the model is not strong enough to accurately select the data points where it is weak. Relying totally on diversity will lead to a challenging and small dataset, making it hard for the model to learn. Depending on density alone will lead to a small subset of similar sentences with uninformative samples causing unnecessary costs. We need hybrid approaches that account for both density and diversity to increase gain in low or very low-resource conditions.

Inspired from the strategy to select in-domain data by \citet{moore2010intelligent}, we present a new hybrid data-driven AL strategy called "Cross-entropy difference". The key idea is to use cross-entropy loss of causal language models (CLM) trained on the labeled and unlabeled data to estimate both diversity and density metrics.

Consider a CLM trained on the unlabelled source data. If a sentence would obtain a smaller cross-entropy loss, it indicates that this sentence is similar to the data distribution of the unlabelled source data. This allows us to measure the density metric and help in selecting sentences that are highly representative. Similarly, higher cross-entropy loss on a language model trained on the labelled source data indicates that the sentence is quite diverse. We use these heuristics and explain how we measure the density and diversity.

 Let the labelled source data be denoted as $\mathcal{L}_{S}$. We train a CLM\footnote{Note that while training a CLM, we initialize with the MLM trained on the monolingual data for better contextualized representations.} on $\mathcal{L}_{S}$ and denote it as $\mathcal{M}_{LS}$. Further, we denote the cross-entropy loss of a sentence $s$ on $\mathcal{M}_{LS}$ as $H(\mathcal{M}_{LS},s)$. We can simply use $H(\mathcal{M}_{LS},s)$ to measure diversity. If the cross-entropy loss is high, than the sentence would score greater in the diversity metric. 

 Recall that the selection strategy scores each sentence in unlabelled source data to estimate its importance. To measure the density metric, we cannot train a language model and evaluate cross-entropy loss on sentences that the model has seen during training. This causes over-fitting and does not provide accurate scores. Therefore, we propose to split the unlabelled source data into two halves and train two separate language models. Then, the first half of the data can be scored using the model trained on the other half and vice-versa.

Let the unlabelled source data be denoted as $\mathcal{U}_{S}$. Due to reasons mentioned above, we split this into two halves $\mathcal{U}_{S1}$ and $\mathcal{U}_{S2}$. We denote the CLM trained on $\mathcal{U}_{S1}$ and $\mathcal{U}_{S2}$ as $\mathcal{M}_{US1}$ and $\mathcal{M}_{US2}$. Now for a sentence $s$ present in $\mathcal{U}_{S1}$, we use $\mathcal{M}_{US2}$ (trained on the other half) to evaluate the cross-entropy loss. Similarly, we use $\mathcal{M}_{US1}$ if $s$ is present in $\mathcal{U}_{S1}$ and estimate the density metric.

Finally, we combine the diversity and density metric using the above cross-entropy losses.  A sentence $s$ in $\mathcal{U}_{S}$ is scored with "Cross-entropy difference" strategy using the following formula:

\begin{multline}
\hspace*{-0.5cm}
    \psi_{\text{ce-diff}}(s) = H(\mathcal{M}_{LS},s) - \mathcal{I}(s \in \mathcal{U}_{S2})\cdot H(\mathcal{M}_{US1},s) \\  - \mathcal{I}(s \in \mathcal{U}_{S1})\cdot H(\mathcal{M}_{US2},s)   
\end{multline}

where $\mathcal{I}(s \in D)$ is $1$ if $s$ is present in $D$ and $0$ otherwise. Higher scores on $H(\mathcal{M}_{LS},s)$ and lower scores on $H(\mathcal{M}_{US},s)$ indicate diversity and density. Therefore, we take the difference of the two to estimate the importance of a sentence.

\section{Experiments and Results}
In this section, we consider English (En) and Kannada (Kn) as our language pair of interest. We chose this as it is truly low-resource, have different writing systems and replicates the challenges faced where AL is needed. We analyze the importance of the proposed techniques to integrate CLWE and evaluate several AL strategies with various annotation budgets.  

\subsection{Datasets}
We assume the availability of monolingual data for the two languages. We use Wikipedia dumps for English and AI4Bharat-IndicNLP corpus \cite{kunchukuttan2020ai4bharat} for Kannada. We chose not to use Wikipedia for Kannada to replicate practical use cases between diverse languages. The parallel data between the languages is from PM-India dataset \cite{haddow2020pmindia}. We train and evaluate according to the split provided by WAT 2021  MultiIndicMT  \cite{nakazawa-etal-2021-overview}. We created our dictionary between English and Kannada using \textit{Kaikki}\footnote{\url{https://kaikki.org/dictionary/Kannada/words.html}}. We discared entries that are not single word-word translations. Statistics about the data are summarized in Table \ref{tab:Dataset}.

\begin{table}[htb]
\resizebox{\columnwidth}{!}{%
\begin{tabular}{|c|c|c|c|c|}
\hline
          &                                                              & \multicolumn{3}{c|}{Total Examples} \\ \hline
Dataset   & Type                                                         & Train      & Valid      & Test      \\ \hline
Wikipedia & Mono (En)                                                    & 46M        & 5K         & 5K        \\ \hline
AI4Bharat & Mono (Kn)                                                    & 15M        & 5K         & 5K        \\ \hline
PMIndia  & \begin{tabular}[c]{@{}c@{}}Parallel\\ (En $\leftrightarrow$ Kn)\end{tabular}   & 29K        & 1.1K       & 2.4K      \\ \hline
Kaikki    & \begin{tabular}[c]{@{}c@{}}Dictionary\\ (En $\leftrightarrow$ Kn)\end{tabular} & 1.1K       & \_         & \_        \\ \hline
\end{tabular}
}
\caption{Overview of the available data.}
\label{tab:Dataset}
\end{table}

\begin{table}[ht]
\resizebox{\columnwidth}{!}{%
\begin{tabular}{@{}ccccc@{}}
\toprule
\multirow{2}{*}{\textit{Word Embedding}} & \multicolumn{2}{c}{BPE}                   & \multicolumn{2}{c}{DP-BPE}         \\ \cmidrule(l){2-5} 
                                         & Kn $\rightarrow$ En & En $\rightarrow$ Kn & Kn $\rightarrow$ En & En $\rightarrow$ Kn \\ \cmidrule(r){1-5}
MWE   & 26.5 & 28.7 & 26.4          & 28.7          \\
CLWE & 25.1 & 27.8 & \textbf{27.3} & \textbf{30.0} \\ \bottomrule
\end{tabular}
}
\caption{Performance of word embeddings v/s pre-processing approach. We report the BLEU scores. Best scores are highlighted in \textbf{bold} for each direction.}
\label{tab:pretrain}
\end{table}

\begin{table*}[ht]
\begin{adjustbox}{width=450pt,center}
\begin{tabular}{@{}ccccccccccccc@{}}
\toprule
\multicolumn{13}{c}{\textit{Selection Strategy}}                                                                                       \\ \midrule
\multirow{2}{*}{\textit{\begin{tabular}[c]{@{}c@{}}Annotation\\ Budget\end{tabular}}} &
  \multicolumn{3}{c}{Random} &
  \multicolumn{3}{c}{RTTL \cite{zeng2019empirical}} &
  \multicolumn{3}{c}{$n$-gram Overlap \cite{eck2005low}} &
  \multicolumn{3}{c}{Cross-entropy diff (\textbf{ours})} \\ \cmidrule(l){2-13} 
 &
  \begin{tabular}[c]{@{}c@{}}No\\ Init\end{tabular} &
  \begin{tabular}[c]{@{}c@{}}UNMT\\ (MWE) Init\end{tabular} &
  \begin{tabular}[c]{@{}c@{}}UNMT \\ (CLWE) Init\end{tabular} &
  \begin{tabular}[c]{@{}c@{}}No\\ Init\end{tabular} &
  \begin{tabular}[c]{@{}c@{}}UNMT\\ (MWE) Init\end{tabular} &
  \begin{tabular}[c]{@{}c@{}}UNMT \\ (CLWE) Init\end{tabular} &
  \begin{tabular}[c]{@{}c@{}}No\\ Init\end{tabular} &
  \begin{tabular}[c]{@{}c@{}}UNMT\\ (MWE) Init\end{tabular} &
  \begin{tabular}[c]{@{}c@{}}UNMT \\ (CLWE) Init\end{tabular} &
  \begin{tabular}[c]{@{}c@{}}No\\ Init\end{tabular} &
  \begin{tabular}[c]{@{}c@{}}UNMT\\ (MWE) Init\end{tabular} &
  \begin{tabular}[c]{@{}c@{}}UNMT \\ (CLWE) Init\end{tabular} \\ \midrule
\multicolumn{13}{c}{Kn $\rightarrow$ En}                                                                                               \\ \midrule
5k  & 7.8  & 16.9 & 18.1 & -           & -  & -            & -  & -  & -  & -           & -           & -            \\
10k & 10.4 & 20.3 & 21.7 & 9.9           & 20.2 & 21.3           & 9.2  & 17.8 & 19.2 & \textbf{10.6} & 20.2             & \textbf{22.2*} \\
15k & 12.8 & 22.4 & 23.1 & 11.3          & 22.0 & \textbf{24.1*}  & 10.2 & 20.2 & 21.2 & 12.0          & 22.1          & \textbf{23.8} \\
20k & 13.3 & 24.3 & 25.2 & \textbf{13.4} & 23.9 & 25.2           & 12.9 & 21.9 & 23.2 & \textbf{13.5} & \textbf{24.4} & \textbf{25.5*} \\ \midrule
\multicolumn{13}{c}{En $\rightarrow$ Kn}                                                                                               \\ \midrule
5k  & 7.6  & 18.7 & 20.3 & -           & -  & -            & -  & -  & -  & -           & -           & -            \\
10k & 11.6 & 22.5 & 24.0 & 11.2          & 22.4 & 23.7           & 10.3 & 20.3 & 21.8 & \textbf{12.2} & 22.5             & \textbf{24.6*} \\
15k & 14.5 & 25.1 & 26.5 & 12.9          & 24.1 & \textbf{26.8*} & 12.7 & 23.1 & 24.4 & 14.5          & 25.0          & 26.5           \\
20k & 15.1 & 26.8 & 27.8 & \textbf{15.9} & 26.5 & \textbf{28.4*} & 15.5 & 24.9 & 26.5 & \textbf{16.0} & \textbf{27.0} & \textbf{28.3}  \\ \bottomrule
\end{tabular}\end{adjustbox}
\caption{Evaluation of AL strategies with respect to different types of pretraining modes and annotation budgets. UNMT (MWE or CLWE) indicates a UNMT model trained using MWE or CLWE while pretraining. We report BLEU scores and append \textbf{*} for the best model given an annotation budget. We highlight in \textbf{bold} if the score is higher than random for that pretraining configuration and budget.}
\label{tab:strategies}
\end{table*}

\subsection{Results on integrating dictionary}
\textbf{What is the benefit of applying DP-BPE and integrating CLWE?}
We evaluate the proposed pretraining approach described to integrate dictionaries in $\S$ \ref{proposedpre}. First, we create monolingual word embeddings (MWE) by joining \textit{fasttext embeddings} for En and Kn and CLWE by mapping the MWE into a common space. Then, we pretrain the models using MWE/CLWE with standard BPE/DP-BPE techniques. Finally, we fine-tune these models on all the parallel data available and report the scores in Table \ref{tab:pretrain}. Comparing these $4$ approaches gives us insight into the role of CLWE and DP-BPE. In the case of "MWE + DP-BPE", we do not have access to dictionary words. However, we simply assume that there is a dictionary and use that for DP-BPE. This tells us if CLWE are necessary. For "CLWE + BPE", the rare words in the dictionary would split into multiple tokens. Therefore, we removed these entries and ended with $390$ word pairs in the dictionary. We mapped the monolingual embeddings with \textit{VecMap} using only these entries. We do this experiment to evaluate the importance of rare words.

We observe similar scores for monolingual embeddings with different type of representations. This shows that the gains from applying DP-BPE are not due to better generalization as in the case of applying dropout in BPE. For CLWE, we find decrease in the performance compared to monolingual embeddings when using standard BPE. We hypothesize this is because of discarding the infrequent words when building CLWE. However, we obtain the best scores by combining CLWE with DP-BPE and gain up to $0.8$ and $1.3$ BLEU in English and Kannada respectively. In this case, we included the rare words in the dictionary when creating our CLWE. This shows that retaining rare words when learning the mapping between embeddings is helpful in exploiting dictionaries for NMT.

\textbf{Do CLWE improve the ability to predict words in the dictionary?} Evaluation metrics like BLEU is not enough to understand the models ability to predict words in the dictionary. We have to also evaluate how many times we predict these words accurately. Therefore, we calculate precision, recall and F1 scores on the dictionary words in the test set. Note that this does not consider the positional information of these words. However, we can judge them together with BLEU. If the model is predicting these words at the wrong positions, then the BLEU scores will be lower. 

We consider two pretraining model configurations: 1) CLWE and DP-BPE (\textbf{With Dict}) 2) MWE with standard BPE (\textbf{No Dict}). Then, we fine-tune these models on different parallel dataset sizes. Finally, we evaluate the models ability to predict English words in the dictionary and report scores in Table \ref{tab:dictwords}.

\begin{table}[ht]
\resizebox{\columnwidth}{!}{%
\begin{tabular}{@{}ccc|cc|cc@{}}
\toprule
 &
  \multicolumn{2}{c|}{\begin{tabular}[c]{@{}c@{}}Precision\\ (\%)\end{tabular}} &
  \multicolumn{2}{c|}{\begin{tabular}[c]{@{}c@{}}Recall\\ (\%)\end{tabular}} &
  \multicolumn{2}{c}{\begin{tabular}[c]{@{}c@{}}F1\\ (\%)\end{tabular}} \\ \midrule
\begin{tabular}[c]{@{}c@{}}Dataset\\ Size\end{tabular} &
  \begin{tabular}[c]{@{}c@{}}No\\ Dict\end{tabular} &
  \begin{tabular}[c]{@{}c@{}}With\\ Dict\end{tabular} &
  \begin{tabular}[c]{@{}c@{}}No\\ Dict\end{tabular} &
  \begin{tabular}[c]{@{}c@{}}With\\ Dict\end{tabular} &
  \begin{tabular}[c]{@{}c@{}}No\\ Dict\end{tabular} &
  \begin{tabular}[c]{@{}c@{}}With\\ Dict\end{tabular} \\ \midrule
10k              & 44.6 & \textbf{49.2} & 48.5          & \textbf{50.0} & 46.5 & \textbf{49.6} \\
15k              & 46.4 & \textbf{48.1} & \textbf{51.6} & 51.3          & 48.9 & \textbf{49.6} \\
20k              & 48.9 & \textbf{49.0} & 51.7          & \textbf{52.5} & 50.3 & \textbf{50.7} \\
Full ($\sim$30k) & 51.0 & \textbf{53.2} & 51.4          & \textbf{55.1} & 52.6 & \textbf{54.1} \\ \bottomrule
\end{tabular}
}
\caption{Impact of CLWE on the test set for predicting English words in the dictionary. We report precision, recall and F1 scores for total $2091$ occurrences. Best scores for each configuration are highlighted in \textbf{bold}.}
\label{tab:dictwords}
\end{table}

We observe that the model's with CLWE are consistently better at predicting these words with relative increase of F1 score up to $3.1\%$. By including rare words in dictionary with help of DP-BPE, we are able to obtain higher performance on these words. Also, the scores in Table \ref{tab:jointbpe} show that including dictionaries with DP-BPE obtain higher BLEU. This indicates the correctness of the predicted positions. However, as we do not explicitly teach the model to predict the dictionary translation \cite{niehues-2021-continuous}, we don't expect significant gains.  

\subsection{Comparision of AL Strategies}

We perform a set of experiments using several AL selection strategies with multiple pretraining configurations. This enables us to assess the role of dictionary in AL and advantages of selection strategies. We consider a batch size of $5k$ and report the scores in Table \ref{tab:strategies}. For the first batch, there is no available labelled data. Therefore, we randomly select $5k$ sentences and initialize our model and labelled data.

Without any initialization, we mostly do not achieve better scores than random with using RTTL or $n$-gram overlap strategy. Our proposed approach Cross-entropy difference is able to beat random most of the time but only with slight gains. Also, the translation quality is not adequate. For pretraining using monolingual embeddings as initialization, we only obtain slight gains than random with our strategy for a budget of $20k$. But, the performance of these models has increased significantly with at least $10$ BLEU. 

For models using pretraining with our proposed approach as initialization, we are consistently able to exploit AL strategies by only spending small amounts on dictionary. Random sampling with a budget of $10k$ and pretraining with monolingual embeddings achieves $20.3$ BLEU when translating to English. While, "Cross-entropy difference" sampling with the same budget but using a small dictionary increases the models performance by $1.9$ BLEU. This shows that building CLWE can be highly beneficial. Furthermore, we observe that the impact of CLWE decreases from around $2$ to $1$ BLEU as we increase the parallel data. Therefore, building CLWE has a bigger impact on very-low resource conditions and might not be as impactful with large amounts of parallel data.

We can conclude that our proposed "Cross-entropy difference" strategy is highly competitive to RTTL in almost all scenarios while RTTL being better in Kannada. However, the "$n$-gram overlap" strategy fails throughout all cases and shows that diversity alone is not a sufficient metric. We need to estimate both density and diversity to gain from data-driven methods for low-annotation budgets. 

\subsection{Impact of freezing the embedding layer} 

\begin{table}[ht]
\resizebox{\columnwidth}{!}{%
\begin{tabular}{@{}ccc@{}}
\toprule
Freezing                 & Kn $\rightarrow$ En & En $\rightarrow$ Kn \\ \midrule
None                     & 25.9                & 27.9                \\
MLM  & 25.0                & 26.9                \\
$\hookrightarrow$ + UNMT & 25.6                & 28.4                \\
$\hookrightarrow$ + Supervised NMT & \textbf{27.3}       & \textbf{30.0}       \\ \bottomrule
\end{tabular}
}
\caption{Analysis on freezing the embedding layer. We report BLEU scores starting from not freezing the embedding layer at any stage and sequentially consider freezing until each phase. $\hookrightarrow$ + UNMT indicates freezing the embeddings at both MLM and UNMT. Best scores are highlighted in \textbf{bold}.}
\label{tab:Freezing}
\end{table}

We proposed to freeze the embedding layer during all stages of training. To understand its role, we evaluate our method with/without freezing at different phases using the full dataset. We report the scores in Table \ref{tab:Freezing}. We observe that freezing at all stages leads to the best performance. By always providing cross-lingual and forcing the model to learn from the CLWE enables the model to exploit them in these low-resource conditions. Also, freezing during only MLM is worse than not freezing at all. We force the model to use CLWE in the pretraining stage and later allow the freedom to alter the embedding layer. We believe this hinders the ability to transfer learning and therefore does not achieve the best results. Moreover, freezing prevents erasing the knowledge from the dictionary and does not allow to drastically change the embeddings weights based on limited parallel data.

\section{Related Work}

\textbf{Low-resource NMT:} Monolingual data being cheaper and available in abundance compared to parallel data, it is widely used to improve NMT models in challenging conditions \cite{haddow2021survey}. One direction is to use pretrained embeddings learned on the monolingual data \cite{di2017monolingual,neishi2017bag}. Another line of work is to pretrain language models on the monolingual data and reload the model for NMT training \cite{ramachandran2017unsupervised,devlin-etal-2019-bert}. Furthermore, several works proposed different variants of back-translation to leverage monolingual data \cite{lample2018unsupervised,guzman2019flores}.

\textbf{Active Learning in NMT:}
There are several works on AL in the context of MT \cite{eck2005low,haffari-etal-2009-active,ambati2012active}. These methods operated and evaluated using phrase-based machine translation systems. Recent works combined AL strategies with the transformer architecture \cite{vaswani2017attention} for NMT systems \cite{zeng2019empirical,zhao2020active}. Interactive NMT includes user feedback to update the NMT model in real-time \cite{peris-casacuberta-2018-active,domingo2019demonstration}. \citet{liu2018learning} uses Deep Imitation Learning to learn the best way to sample using a high resource language pair. Our work differs in terms of analysis with respect to the annotation budget and investigate ways to leverage monolingual data to reduce costs with AL.

\section{Conclusion}
The main goal of the paper was to design a high-quality NMT system with limited annotation costs. To achieve this, we designed a cost-effective training procedure by proposing improvements in multiple avenues. First, we showed the necessity of pretraining with monolingual data. This is useful as the monolingual data does not require any labelling and improves the models significantly. Moreover, it enables us to gain from selection strategies. Second, we suggested a pretraining procedure by integrating a dictionary which can be created cheaply. We proposed DP-BPE to include the rare words in the dictionary while learning the alignment. Further, we showed the importance of including these rare words from our experiments. Using our approach, we were able to increase the models ability to predict these words. Finally, we presented a novel data-driven strategy "Cross-entropy difference" that is helpful in low-resource scenarios. We empirically showed that sampling using our strategy achieves better scores than random consistently and is competitive to the SOTA approach RTTL.   

Pretraining with auxiliary data of similar high-resource languages can substantially increase the model's quality. Building such multilingual models can greatly increase the potential of model-driven strategies. Also, designing AL strategies for constructing a dictionary can even further decrease costs while increasing gains. We leave these directions as future work.

\bibliography{anthology,custom}

\begin{thebibliography}{41}
\expandafter\ifx\csname natexlab\endcsname\relax\def\natexlab#1{#1}\fi

\bibitem[{Ambati(2012)}]{ambati2012active}
Vamshi Ambati. 2012.
\newblock \href {https://apps.dtic.mil/sti/pdfs/AD1144428.pdf} {\emph{Active
  learning and crowdsourcing for machine translation in low resource
  scenarios}}.
\newblock Ph.D. thesis, Carnegie Mellon University.

\bibitem[{Artetxe et~al.(2017)Artetxe, Labaka, and
  Agirre}]{artetxe2017learning}
Mikel Artetxe, Gorka Labaka, and Eneko Agirre. 2017.
\newblock \href {https://aclanthology.org/P17-1042/} {Learning bilingual word
  embeddings with (almost) no bilingual data}.
\newblock In \emph{Proceedings of the 55th Annual Meeting of the Association
  for Computational Linguistics (Volume 1: Long Papers)}, pages 451--462.

\bibitem[{Artetxe et~al.(2018)Artetxe, Labaka, Agirre, and
  Cho}]{artetxe2018unsupervised}
Mikel Artetxe, Gorka Labaka, Eneko Agirre, and Kyunghyun Cho. 2018.
\newblock \href {https://arxiv.org/pdf/1710.11041.pdf} {Unsupervised neural
  machine translation}.
\newblock In \emph{International Conference on Learning Representations}.

\bibitem[{Bahdanau et~al.(2015)Bahdanau, Cho, and Bengio}]{bahdanau2015neural}
Dzmitry Bahdanau, Kyung~Hyun Cho, and Yoshua Bengio. 2015.
\newblock \href {https://arxiv.org/pdf/1409.0473} {Neural machine translation
  by jointly learning to align and translate}.
\newblock In \emph{3rd International Conference on Learning Representations,
  ICLR 2015}.

\bibitem[{Banerjee et~al.(2021)Banerjee, V~Murthy, and
  Bhattacharya}]{banerjee2021crosslingual}
Tamali Banerjee, Rudra V~Murthy, and Pushpak Bhattacharya. 2021.
\newblock \href {https://aclanthology.org/2021.mtsummit-research.3}
  {Crosslingual embeddings are essential in {UNMT} for distant languages: An
  {E}nglish to {I}ndo{A}ryan case study}.
\newblock In \emph{Proceedings of Machine Translation Summit XVIII: Research
  Track}, pages 23--34, Virtual. Association for Machine Translation in the
  Americas.

\bibitem[{Bojanowski et~al.(2017)Bojanowski, Grave, Joulin, and
  Mikolov}]{bojanowski2017enriching}
Piotr Bojanowski, Edouard Grave, Armand Joulin, and Tomas Mikolov. 2017.
\newblock \href {https://doi.org/10.1162/tacl_a_00051} {Enriching word vectors
  with subword information}.
\newblock \emph{Transactions of the Association for Computational Linguistics},
  5:135--146.

\bibitem[{Chronopoulou et~al.(2021)Chronopoulou, Stojanovski, and
  Fraser}]{chronopoulou-etal-2021-improving}
Alexandra Chronopoulou, Dario Stojanovski, and Alexander Fraser. 2021.
\newblock \href {https://doi.org/10.18653/v1/2021.naacl-main.16} {Improving the
  lexical ability of pretrained language models for unsupervised neural machine
  translation}.
\newblock In \emph{Proceedings of the 2021 Conference of the North American
  Chapter of the Association for Computational Linguistics: Human Language
  Technologies}, pages 173--180, Online. Association for Computational
  Linguistics.

\bibitem[{Conneau and Lample(2019)}]{conneau2019cross}
Alexis Conneau and Guillaume Lample. 2019.
\newblock \href
  {https://proceedings.neurips.cc/paper/2019/file/c04c19c2c2474dbf5f7ac4372c5b9af1-Paper.pdf}
  {Cross-lingual language model pretraining}.
\newblock \emph{Advances in Neural Information Processing Systems},
  32:7059--7069.

\bibitem[{Devlin et~al.(2019)Devlin, Chang, Lee, and
  Toutanova}]{devlin-etal-2019-bert}
Jacob Devlin, Ming-Wei Chang, Kenton Lee, and Kristina Toutanova. 2019.
\newblock \href {https://doi.org/10.18653/v1/N19-1423} {{BERT}: Pre-training of
  deep bidirectional transformers for language understanding}.
\newblock In \emph{Proceedings of the 2019 Conference of the North {A}merican
  Chapter of the Association for Computational Linguistics: Human Language
  Technologies, Volume 1 (Long and Short Papers)}, pages 4171--4186,
  Minneapolis, Minnesota. Association for Computational Linguistics.

\bibitem[{Di~Gangi and Federico(2017)}]{di2017monolingual}
Mattia~Antonino Di~Gangi and Marcello Federico. 2017.
\newblock \href
  {https://www.researchgate.net/profile/Mattia-Di-Gangi/publication/321706277_Monolingual_Embeddings_for_Low_Resourced_Neural_Machine_Translation/links/5a2c083945851552ae7c79b5/Monolingual-Embeddings-for-Low-Resourced-Neural-Machine-Translation.pdf}
  {Monolingual embeddings for low resourced neural machine translation}.
\newblock In \emph{Proceedings of the 14th international workshop on spoken
  language translation (IWSLT’17)}, pages 97--104.

\bibitem[{Domingo et~al.(2019)Domingo, Garc{\'\i}a-Mart{\'\i}nez, Estela,
  Bi{\'e}, Helle, Peris, Casacuberta, and Herranz}]{domingo2019demonstration}
Miguel Domingo, Mercedes Garc{\'\i}a-Mart{\'\i}nez, Amando Estela, Laurent
  Bi{\'e}, Alexandre Helle, {\'A}lvaro Peris, Francisco Casacuberta, and
  Manuerl Herranz. 2019.
\newblock \href {https://arxiv.org/abs/1906.09000} {Demonstration of a neural
  machine translation system with online learning for translators}.
\newblock \emph{arXiv preprint arXiv:1906.09000}.

\bibitem[{Eck et~al.(2005)Eck, Vogel, and Waibel}]{eck2005low}
Matthias Eck, Stephan Vogel, and Alex Waibel. 2005.
\newblock \href {https://aclanthology.org/2005.iwslt-1.7/} {Low cost
  portability for statistical machine translation based on n-gram frequency and
  tf-idf}.
\newblock In \emph{International Workshop on Spoken Language Translation
  (IWSLT) 2005}.

\bibitem[{Fraser(2020)}]{fraser-2020-findings}
Alexander Fraser. 2020.
\newblock \href {https://aclanthology.org/2020.wmt-1.80} {Findings of the {WMT}
  2020 shared tasks in unsupervised {MT} and very low resource supervised
  {MT}}.
\newblock In \emph{Proceedings of the Fifth Conference on Machine Translation},
  pages 765--771, Online. Association for Computational Linguistics.

\bibitem[{Garcia et~al.(2021)Garcia, Siddhant, Firat, and
  Parikh}]{garcia2021harnessing}
Xavier Garcia, Aditya Siddhant, Orhan Firat, and Ankur Parikh. 2021.
\newblock \href {https://aclanthology.org/2021.naacl-main.89/} {Harnessing
  multilinguality in unsupervised machine translation for rare languages}.
\newblock In \emph{Proceedings of the 2021 Conference of the North American
  Chapter of the Association for Computational Linguistics: Human Language
  Technologies}, pages 1126--1137.

\bibitem[{Guzm{\'a}n et~al.(2019)Guzm{\'a}n, Chen, Ott, Pino, Lample, Koehn,
  Chaudhary, and Ranzato}]{guzman2019flores}
Francisco Guzm{\'a}n, Peng-Jen Chen, Myle Ott, Juan Pino, Guillaume Lample,
  Philipp Koehn, Vishrav Chaudhary, and Marc’Aurelio Ranzato. 2019.
\newblock The flores evaluation datasets for low-resource machine translation:
  Nepali--english and sinhala--english.
\newblock In \emph{Proceedings of the 2019 Conference on Empirical Methods in
  Natural Language Processing and the 9th International Joint Conference on
  Natural Language Processing (EMNLP-IJCNLP)}, pages 6098--6111.

\bibitem[{Haddow et~al.(2021)Haddow, Bawden, Barone, Helcl, and
  Birch}]{haddow2021survey}
Barry Haddow, Rachel Bawden, Antonio Valerio~Miceli Barone, Jind{\v{r}}ich
  Helcl, and Alexandra Birch. 2021.
\newblock \href {https://arxiv.org/abs/2109.00486} {Survey of low-resource
  machine translation}.
\newblock \emph{arXiv preprint arXiv:2109.00486}.

\bibitem[{Haddow and Kirefu(2020)}]{haddow2020pmindia}
Barry Haddow and Faheem Kirefu. 2020.
\newblock \href {https://arxiv.org/abs/2001.09907} {{PMIndia}--a collection of
  parallel corpora of languages of india}.
\newblock \emph{arXiv preprint arXiv:2001.09907}.

\bibitem[{Haffari et~al.(2009)Haffari, Roy, and
  Sarkar}]{haffari-etal-2009-active}
Gholamreza Haffari, Maxim Roy, and Anoop Sarkar. 2009.
\newblock \href {https://aclanthology.org/N09-1047} {Active learning for
  statistical phrase-based machine translation}.
\newblock In \emph{Proceedings of Human Language Technologies: The 2009 Annual
  Conference of the North {A}merican Chapter of the Association for
  Computational Linguistics}, pages 415--423, Boulder, Colorado. Association
  for Computational Linguistics.

\bibitem[{Kim et~al.(2020)Kim, Gra{\c{c}}a, and Ney}]{kim2020and}
Yunsu Kim, Miguel Gra{\c{c}}a, and Hermann Ney. 2020.
\newblock \href {https://aclanthology.org/2020.eamt-1.5} {When and why is
  unsupervised neural machine translation useless?}
\newblock In \emph{Proceedings of the 22nd Annual Conference of the European
  Association for Machine Translation}, pages 35--44.

\bibitem[{Koehn et~al.(2007)Koehn, Hoang, Birch, Callison-Burch, Federico,
  Bertoldi, Cowan, Shen, Moran, Zens et~al.}]{koehn2007moses}
Philipp Koehn, Hieu Hoang, Alexandra Birch, Chris Callison-Burch, Marcello
  Federico, Nicola Bertoldi, Brooke Cowan, Wade Shen, Christine Moran, Richard
  Zens, et~al. 2007.
\newblock \href {https://aclanthology.org/P07-2045/} {Moses: Open source
  toolkit for statistical machine translation}.
\newblock In \emph{Proceedings of the 45th annual meeting of the ACL on
  interactive poster and demonstration sessions}, pages 177--180. Association
  for Computational Linguistics.

\bibitem[{Koneru et~al.(2021)Koneru, Liu, and Niehues}]{koneru2021unsupervised}
Sai Koneru, Danni Liu, and Jan Niehues. 2021.
\newblock Unsupervised machine translation on dravidian languages.
\newblock In \emph{Proceedings of the First Workshop on Speech and Language
  Technologies for Dravidian Languages}, pages 55--64.

\bibitem[{Kunchukuttan et~al.(2020)Kunchukuttan, Kakwani, Golla, Bhattacharyya,
  Khapra, Kumar et~al.}]{kunchukuttan2020ai4bharat}
Anoop Kunchukuttan, Divyanshu Kakwani, Satish Golla, Avik Bhattacharyya,
  Mitesh~M Khapra, Pratyush Kumar, et~al. 2020.
\newblock \href {https://arxiv.org/pdf/2005.00085.pdf} {Ai4bharat-indicnlp
  corpus: Monolingual corpora and word embeddings for indic languages}.
\newblock \emph{arXiv preprint arXiv:2005.00085}.

\bibitem[{Lample et~al.(2018)Lample, Conneau, Denoyer, and
  Ranzato}]{lample2018unsupervised}
Guillaume Lample, Alexis Conneau, Ludovic Denoyer, and Marc'Aurelio Ranzato.
  2018.
\newblock \href {https://openreview.net/pdf?id=rkYTTf-AZ} {Unsupervised machine
  translation using monolingual corpora only}.
\newblock In \emph{International Conference on Learning Representations}.

\bibitem[{Li et~al.(2020)Li, Zhao, Wang, Utiyama, and
  Sumita}]{li-etal-2020-reference}
Zuchao Li, Hai Zhao, Rui Wang, Masao Utiyama, and Eiichiro Sumita. 2020.
\newblock \href {https://doi.org/10.18653/v1/2020.findings-emnlp.371}
  {Reference language based unsupervised neural machine translation}.
\newblock In \emph{Findings of the Association for Computational Linguistics:
  EMNLP 2020}, pages 4151--4162, Online. Association for Computational
  Linguistics.

\bibitem[{Liu et~al.(2018)Liu, Buntine, and Haffari}]{liu2018learning}
Ming Liu, Wray Buntine, and Gholamreza Haffari. 2018.
\newblock \href {https://doi.org/10.18653/v1/K18-1033} {Learning to actively
  learn neural machine translation}.
\newblock In \emph{Proceedings of the 22nd Conference on Computational Natural
  Language Learning}, pages 334--344, Brussels, Belgium. Association for
  Computational Linguistics.

\bibitem[{Liu et~al.(2020)Liu, Gu, Goyal, Li, Edunov, Ghazvininejad, Lewis, and
  Zettlemoyer}]{liu2020multilingual}
Yinhan Liu, Jiatao Gu, Naman Goyal, Xian Li, Sergey Edunov, Marjan
  Ghazvininejad, Mike Lewis, and Luke Zettlemoyer. 2020.
\newblock \href {https://arxiv.org/pdf/2001.08210.pdf} {Multilingual denoising
  pre-training for neural machine translation}.
\newblock \emph{Transactions of the Association for Computational Linguistics},
  8:726--742.

\bibitem[{Marchisio et~al.(2020)Marchisio, Duh, and Koehn}]{marchisio2020does}
Kelly Marchisio, Kevin Duh, and Philipp Koehn. 2020.
\newblock \href {https://aclanthology.org/2020.wmt-1.68/} {When does
  unsupervised machine translation work?}
\newblock In \emph{Proceedings of the Fifth Conference on Machine Translation},
  pages 571--583.

\bibitem[{Moore and Lewis(2010)}]{moore2010intelligent}
Robert~C Moore and William Lewis. 2010.
\newblock \href {https://aclanthology.org/P10-2041/} {Intelligent selection of
  language model training data}.
\newblock In \emph{Proceedings of the ACL 2010 Conference Short Papers}, pages
  220--224.

\bibitem[{Nakazawa et~al.(2021)Nakazawa, Nakayama, Ding, Dabre, Higashiyama,
  Mino, Goto, Pa~Pa, Kunchukuttan, Parida, Bojar, Chu, Eriguchi, Abe, Oda, and
  Kurohashi}]{nakazawa-etal-2021-overview}
Toshiaki Nakazawa, Hideki Nakayama, Chenchen Ding, Raj Dabre, Shohei
  Higashiyama, Hideya Mino, Isao Goto, Win Pa~Pa, Anoop Kunchukuttan,
  Shantipriya Parida, Ond{\v{r}}ej Bojar, Chenhui Chu, Akiko Eriguchi, Kaori
  Abe, Yusuke Oda, and Sadao Kurohashi. 2021.
\newblock \href {https://doi.org/10.18653/v1/2021.wat-1.1} {Overview of the 8th
  workshop on {A}sian translation}.
\newblock In \emph{Proceedings of the 8th Workshop on Asian Translation
  (WAT2021)}, pages 1--45, Online. Association for Computational Linguistics.

\bibitem[{Neishi et~al.(2017)Neishi, Sakuma, Tohda, Ishiwatari, Yoshinaga, and
  Toyoda}]{neishi2017bag}
Masato Neishi, Jin Sakuma, Satoshi Tohda, Shonosuke Ishiwatari, Naoki
  Yoshinaga, and Masashi Toyoda. 2017.
\newblock A bag of useful tricks for practical neural machine translation:
  Embedding layer initialization and large batch size.
\newblock In \emph{Proceedings of the 4th Workshop on Asian Translation
  (WAT2017)}, pages 99--109.

\bibitem[{Niehues(2021)}]{niehues-2021-continuous}
Jan Niehues. 2021.
\newblock \href {https://doi.org/10.18653/v1/2021.eacl-main.70} {Continuous
  learning in neural machine translation using bilingual dictionaries}.
\newblock In \emph{Proceedings of the 16th Conference of the European Chapter
  of the Association for Computational Linguistics: Main Volume}, pages
  830--840, Online. Association for Computational Linguistics.

\bibitem[{Papineni et~al.(2002)Papineni, Roukos, Ward, and
  Zhu}]{papineni2002bleu}
Kishore Papineni, Salim Roukos, Todd Ward, and Wei-Jing Zhu. 2002.
\newblock \href {https://aclanthology.org/P02-1040/} {Bleu: a method for
  automatic evaluation of machine translation}.
\newblock In \emph{Proceedings of the 40th annual meeting of the Association
  for Computational Linguistics}, pages 311--318.

\bibitem[{Peris and Casacuberta(2018)}]{peris-casacuberta-2018-active}
{\'A}lvaro Peris and Francisco Casacuberta. 2018.
\newblock \href {https://doi.org/10.18653/v1/K18-1015} {Active learning for
  interactive neural machine translation of data streams}.
\newblock In \emph{Proceedings of the 22nd Conference on Computational Natural
  Language Learning}, pages 151--160, Brussels, Belgium. Association for
  Computational Linguistics.

\bibitem[{Post(2018)}]{post2018call}
Matt Post. 2018.
\newblock \href {https://aclanthology.org/W18-6319/} {A call for clarity in
  reporting bleu scores}.
\newblock In \emph{Proceedings of the Third Conference on Machine Translation:
  Research Papers}, pages 186--191.

\bibitem[{Ramachandran et~al.(2017)Ramachandran, Liu, and
  Le}]{ramachandran2017unsupervised}
Prajit Ramachandran, Peter~J Liu, and Quoc Le. 2017.
\newblock \href {https://aclanthology.org/D17-1039.pdf} {Unsupervised
  pretraining for sequence to sequence learning}.
\newblock In \emph{Proceedings of the 2017 Conference on Empirical Methods in
  Natural Language Processing}, pages 383--391.

\bibitem[{Sennrich et~al.(2016{\natexlab{a}})Sennrich, Haddow, and
  Birch}]{sennrich2016improving}
Rico Sennrich, Barry Haddow, and Alexandra Birch. 2016{\natexlab{a}}.
\newblock \href {https://aclanthology.org/P16-1009/} {Improving neural machine
  translation models with monolingual data}.
\newblock In \emph{54th Annual Meeting of the Association for Computational
  Linguistics}, pages 86--96. Association for Computational Linguistics (ACL).

\bibitem[{Sennrich et~al.(2016{\natexlab{b}})Sennrich, Haddow, and
  Birch}]{sennrich2016neural}
Rico Sennrich, Barry Haddow, and Alexandra Birch. 2016{\natexlab{b}}.
\newblock \href {https://aclanthology.org/P16-1162/} {Neural machine
  translation of rare words with subword units}.
\newblock In \emph{Proceedings of the 54th Annual Meeting of the Association
  for Computational Linguistics (Volume 1: Long Papers)}, pages 1715--1725.

\bibitem[{Vaswani et~al.(2017)Vaswani, Shazeer, Parmar, Uszkoreit, Jones,
  Gomez, Kaiser, and Polosukhin}]{vaswani2017attention}
Ashish Vaswani, Noam Shazeer, Niki Parmar, Jakob Uszkoreit, Llion Jones,
  Aidan~N Gomez, {\L}ukasz Kaiser, and Illia Polosukhin. 2017.
\newblock \href
  {https://papers.nips.cc/paper/2017/hash/3f5ee243547dee91fbd053c1c4a845aa-Abstract.html}
  {Attention is all you need}.
\newblock In \emph{Advances in neural information processing systems}, pages
  5998--6008.

\bibitem[{Vincent et~al.(2008)Vincent, Larochelle, Bengio, and
  Manzagol}]{vincent2008extracting}
Pascal Vincent, Hugo Larochelle, Yoshua Bengio, and Pierre-Antoine Manzagol.
  2008.
\newblock \href {https://dl.acm.org/doi/10.1145/1390156.1390294} {Extracting
  and composing robust features with denoising autoencoders}.
\newblock In \emph{Proceedings of the 25th international conference on Machine
  learning}, pages 1096--1103.

\bibitem[{Zeng et~al.(2019)Zeng, Garg, Chatterjee, Nallasamy, and
  Paulik}]{zeng2019empirical}
Xiangkai Zeng, Sarthak Garg, Rajen Chatterjee, Udhyakumar Nallasamy, and
  Matthias Paulik. 2019.
\newblock \href {https://aclanthology.org/D19-6110/} {Empirical evaluation of
  active learning techniques for neural mt}.
\newblock In \emph{Proceedings of the 2nd Workshop on Deep Learning Approaches
  for Low-Resource NLP (DeepLo 2019)}, pages 84--93.

\bibitem[{Zhao et~al.(2020)Zhao, Zhang, Zhou, and Zhang}]{zhao2020active}
Yuekai Zhao, Haoran Zhang, Shuchang Zhou, and Zhihua Zhang. 2020.
\newblock \href {https://aclanthology.org/2020.findings-emnlp.162.pdf} {Active
  learning approaches to enhancing neural machine translation: An empirical
  study}.
\newblock In \emph{Proceedings of the 2020 Conference on Empirical Methods in
  Natural Language Processing: Findings}, pages 1796--1806.

\end{thebibliography}
\bibliographystyle{acl_natbib}

\appendix
\section{Appendix}

\subsection{AL Framework}

\begin{algorithm}
\caption{General AL Algorithm}
\begin{algorithmic}
\REQUIRE Parallel Data $\mathcal{D}_{P}$, \\ Monolingual Data $\mathcal{D}_{M}$,\\ Unlabelled in-domain source data $\mathcal{U}_{S}$, \\ Batch size $\mathcal{B}$, Selection strategy $\psi()$
\STATE $\mathcal{M}_{PRE} \leftarrow PRETRAIN(\mathcal{D}_{M},empty)$;
\STATE $\mathcal{M} \leftarrow SNMT(\mathcal{D}_{P},\mathcal{M}_{PRE})$;
\WHILE{$Budget \neq 0$}
\FOR{$x \in \mathcal{U}_{S}$}
\STATE $f(x) \mathrel{+}= \psi(x,\mathcal{U}_{S},\mathcal{D}_{P},\mathcal{M})$;
\ENDFOR
\STATE $X_{B} = Topscoring(f(x),\mathcal{B})$;
\STATE $Y_{B} = HumanTranslated(X_{B})$;
\STATE $\mathcal{U}_{S} = \mathcal{U}_{S} - X_{B}$;
\STATE $\mathcal{D}_{P} = \mathcal{D}_{P} \cup (X_{B},Y_{B})$;
\STATE $\mathcal{M} \leftarrow SNMT(\mathcal{D}_{P},\mathcal{M}_{PRE})$;
\ENDWHILE
\RETURN $\mathcal{M},\mathcal{D}_{P}$
\end{algorithmic}
\end{algorithm}

\subsection{Pre-processing and Hyperparameters}
We tokenize the data with Moses \cite{koehn2007moses} for English and \textit{Indic-NLP-Library}\footnote{\url{https://github.com/anoopkunchukuttan/indic_nlp_library}} for Kannada. We learn sub-words using BPE \cite{sennrich2016neural} with $50k$ merge operations on concatenating subset of English and Kannada data. We report detokenized BLEU \cite{papineni2002bleu} using SacreBLEU\footnote{BLEU+case.mixed+numrefs.1+smooth.exp+tok.
spm+version.1.4.12} \cite{post2018call}. We use the SOTA Transformer architecture \cite{vaswani2017attention} for building NMT models. For pretraining, we use a transformer with $6$ layers and $8$ heads and an embedding dimension of $1024$. While fine-tuning on the parallel data, we use label-smoothing of $0.2$, activation dropout of $0.2$ and attention dropout of $0.2$ as we have limited data. The language models for the "Cross-entropy difference" strategy use the pretrained MLM model as initialization before training on the CLM objective. For the models that do not use any initialization in Table \ref{tab:strategies}, we use a smaller model with $5$ layers and $2$ heads and an embedding dimesion of $512$. We use the same value for the regularization parameters as mentioned in the pretraining architecture. Furthermore, the CLM for the "Cross-entropy difference" strategy use a transformer with $3$ layers and $2$ heads as there is no pretrained model. We use the \textit{XLM}\footnote{\url{https://github.com/facebookresearch/XLM}} code base to perform our experiments and set the other parameters to default.

\end{document}